\documentclass[letterpaper]{article} % DO NOT CHANGE THIS
\usepackage{aaai2026}  % DO NOT CHANGE THIS
\usepackage{times}  % DO NOT CHANGE THIS
\usepackage{helvet}  % DO NOT CHANGE THIS
\usepackage{courier}  % DO NOT CHANGE THIS
\usepackage[hyphens]{url}  % DO NOT CHANGE THIS
\usepackage{graphicx} % DO NOT CHANGE THIS
\usepackage{natbib}  % DO NOT CHANGE THIS AND DO NOT ADD ANY OPTIONS TO IT
\usepackage{caption} % DO NOT CHANGE THIS AND DO NOT ADD ANY OPTIONS TO IT
\usepackage{cite}
\usepackage{lscape} 
\usepackage{mathtools}
\usepackage{enumitem}
\usepackage{multirow}
\usepackage{amsthm}
\usepackage{amssymb}
\usepackage{amsmath}
\usepackage{blindtext}
\usepackage{algpseudocode}

\theoremstyle{plain}% Theorem-like structures provided by amsthm.sty

\theoremstyle{definition}

\theoremstyle{remark}

\usepackage{graphicx}
\usepackage{subcaption}
\usepackage{xspace}
\usepackage{algorithm}
\usepackage{algpseudocode}
\usepackage{multirow}
\usepackage{booktabs}
\usepackage{makecell}
\usepackage{comment}
\usepackage[inkscapelatex=false]{svg}
\usepackage{xcolor}
\usepackage{url}
\usepackage{pifont}
\usepackage{tablefootnote}
\usepackage{threeparttable}
\usepackage{tcolorbox}
\newcommand{\rqq}[1]{
\begin{center}
\begin{tcolorbox}[width=\columnwidth, 
 colback=gray!5!white, 
 colframe=cyan!60!black, 
boxrule=0.5px,
left=2pt,
right=2pt,
top=2pt,
bottom=2pt,
arc=5pt,
auto outer arc]
{#1}
\end{tcolorbox}
\end{center}
}

\title{Certified but Fooled! Breaking Certified Defences with Ghost Certificates}
\author {
    Quoc Viet Vo\textsuperscript{\rm 1},
    Tashreque M. Haq\textsuperscript{\rm 1},
    Paul Montague\textsuperscript{\rm 3},
    Tamas Abraham\textsuperscript{\rm 3},
    Ehsan Abbasnejad\textsuperscript{\rm 2},
    Damith C. Ranasinghe\textsuperscript{\rm 1}
}
\affiliations {
    \textsuperscript{\rm 1}University of Adelaide\\
    \textsuperscript{\rm 2}Monash University\\
    \textsuperscript{\rm 3}Defence Science and Technology Group\\
    quocviet.vo@adelaide.edu.au, tashrequemohammed.haq@adelaide.edu.au, paul.montague@defence.gov.au, tamas.abraham@defence.gov.au, ehsan.abbasnejad@monash.edu, damith.ranasinghe@adelaide.edu.au
}

\usepackage{amsmath,amsfonts,bm}

\def\eqref#1{equation~\ref{#1}}
\def\1{\bm{1}}

\DeclareMathAlphabet{\mathsfit}{\encodingdefault}{\sfdefault}{m}{sl}
\SetMathAlphabet{\mathsfit}{bold}{\encodingdefault}{\sfdefault}{bx}{n}

\newcommand{\name}[0]{{\textsf{GhostCert}}}
\newcommand{\shadow}{{\small\textsf{Shadow Attack}}\xspace}

\begin{document}

\maketitle

\begin{abstract}
Certified defenses promise provable robustness guarantees. We study the malicious exploitation of probabilistic certification frameworks to  better understand the limits of guarantee provisions. Now, the objective is to not only mislead a classifier, but also manipulate the certification process to generate a robustness guarantee for an adversarial input---\textit{certificate spoofing}. A recent study in ICLR demonstrated that crafting large perturbations can shift inputs far into regions capable of generating a certificate for an incorrect class. Our study investigates if perturbations needed to cause a misclassification and yet coax a certified model into issuing a deceptive, large robustness radius for a target class can still be  made \textit{small} and \textit{imperceptible}. We explore the idea of region-focused adversarial examples to craft imperceptible perturbations, spoof certificates and achieve certification radii larger than the source class---\textit{ghost certificates}. Extensive evaluations with the \texttt{ImageNet} demonstrate the ability to effectively bypass state-of-the-art certified defenses such as Densepure. Our work underscores the need to better understand the limits of robustness certification methods. 
\end{abstract}
\vspace{-1mm}

\section{Introduction}
\label{sec:introduction}
Deep neural networks (DNNs) are vulnerable to adversarial examples---carefully crafted perturbations to manipulate inputs to coerce incorrect model decisions whilst remaining imperceptible to human observers~\citep{biggio2013evasion, Szegedy2013, papernot2017, Carlini2017, Madry2017, Brendel2018}. In response, various empirical defenses such as adversarial training and preprocessing inputs are proposed~\cite{dalvi2004adversarialperturbation, papernot2016distillation, buckman2018thermometer, guo2018countering, samangouei2018defensegan, Shafahi2019, wong2020fastbetterfreerevisiting, rebuffi2021fixingdataaugmentationimprove,Doan2022}. But, demonstrating empirical robustness alone does not facilitate reasoning about robustness guarantees, and strong adaptive attacks can bypass defenses~\cite{carlini2017adversarialexampleseasilydetected, Athalye2018, Tramer2020, croce2020reliableevaluation, wong2020fastbetterfreerevisiting}. So, certified robustness has emerged to provide provable lower bounds on model accuracy under bounded perturbations.

Conceptually, certification methods provide both a \textit{model} for a task to generate a prediction and a \textit{verifier} for generating a certificate guaranteeing an input image is not an adversarial example under a predefined threat model. Existing certification methods in the vision domain predominantly focus on $l_2$ or $l_\infty$-bounded threat models, effectively ensuring that all inputs within an $\epsilon$-ball neighbourhood of a given image are consistently classified under the same label. We focus on the probabilistic \textit{Randomized Smoothing}~\cite{cohen2019certified, zhang2019stableandefficienttraining} frameworks offering scalable certification for tasks.

\begin{figure*}[t]
    \centering
    % \includesvg[width=0.83\textwidth]{spoof-figures/fig1_02_08.pdf}
    \includegraphics[width=0.83\textwidth]{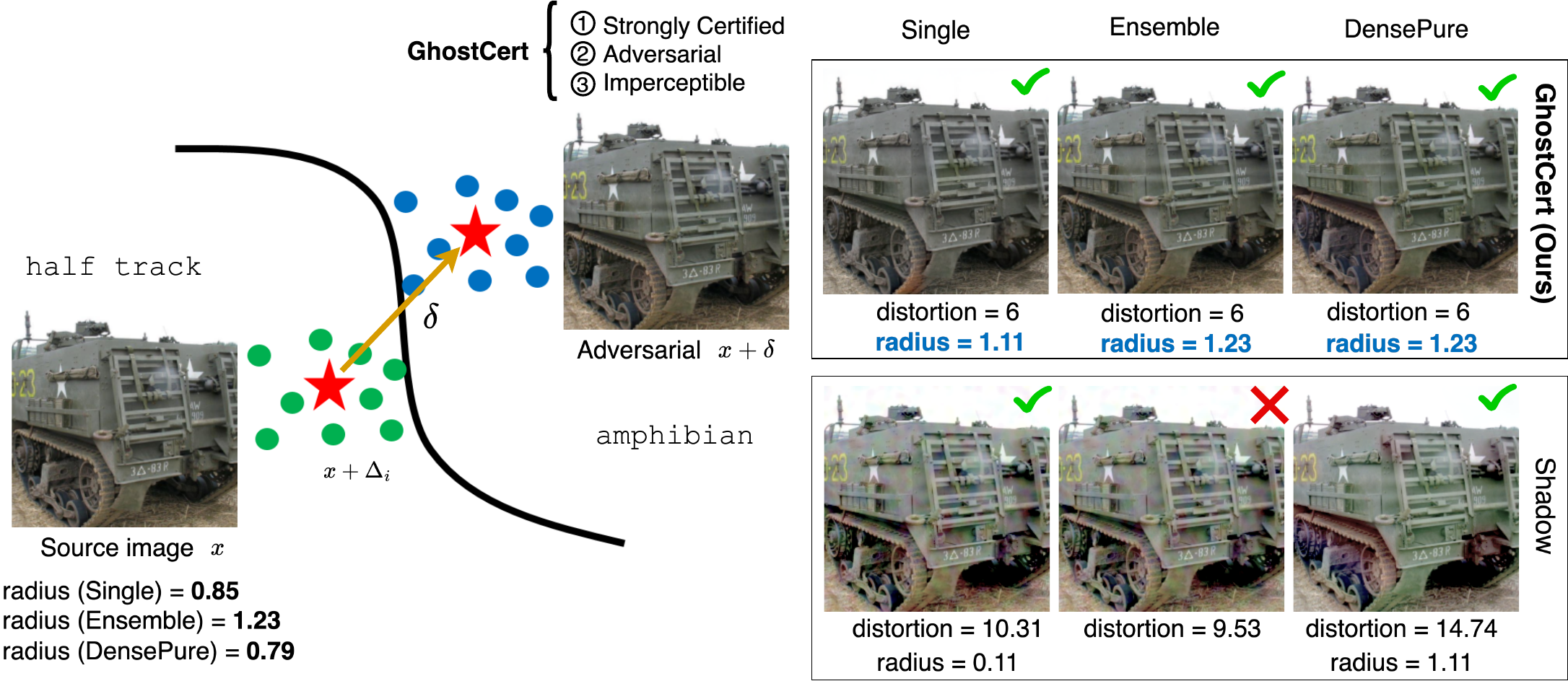}
    \caption{Overview of our attack formulation. For a given source image (\texttt{half track}) and certification radii, we show the corresponding adversarial examples created by our attack \name{} and \shadow{} in ICLR~\cite{Ghiasi2020} against three certified defense methods: Randomized Smoothing (with Resnet50), Smoothed Ensemble, and DensePure (Diffusion based denoiser \& Transformer under Randomized Smoothing). Shadow fails to generate a spoofed certificate (\color{red}\ding{55}\color{black}) for Smoothed Ensemble even with a \textit{larger} distortion. \name{} generates more \textit{natural-looking} adversarials across all three defenses while achieving misclassification with: i)~higher spoofed certification radii; and ii)~significantly lower $l_2$ norms ($||\delta||_2$) compared to the \shadow (\textit{Adversarial} and \textit{Imperceptible}). \name{} results also surpass the certification radii of the source image (\textit{Strongly Certified})---see Fig.~\ref{fig:naturalism_score} for results of a user-study on imperceptibility. \textbf{Code}: \color{blue}\underline{\url{https://github.com/ghostcert}}\color{black}}
    \label{fig:attack_illustration}
    \vspace{-5mm}
\end{figure*}

Unlike conventional attacks---eg., PGD~\citep{Madry2017}, Wasserstein~\citep{Levine2020}, and semantic attacks~\citep{Shahin2020, Bhattad2020}---we explore the adversarial exploitation of certified models that \textit{both} mislead the classifier and falsely receive robustness certificates. If attackers can generate an adversarial example that remains imperceptible and semantically consistent, while ensuring neighbouring images within a certification radius share an incorrect label---a verifier is fooled with a spoofed certificate~\cite{Ghiasi2020}. 

\vspace{1mm}
\noindent\textbf{Our Study.~}Spoofing attacks introduce a new dimension to the adversarial threat landscape by targeting \textit{both} the classification and certification processes. To investigate the assurances provided by certified defences, we introduce a new certificate spoofing  mechanism that systematically undermines classification and certification mechanisms.  
\vspace{-1mm}
\rqq{
Our idea is to explore \textit{if} region-based manipulation of inputs to constrain translation in the input-space---a means to preserve input semantics whilst minimizing manipulation---can still shift inputs in the latent-space just far enough to achieve a misclassified label from a model and yet, a \textit{large} radius certificate from a verifier.
}
\vspace{-1mm}
To achieve our idea, we construct a malicious objective to induce high-probability of misclassified neighborhoods to yield certified radii—beyond just fooling classifiers.

\vspace{1mm}
\noindent\textbf{Contributions and Findings~}
\begin{itemize}
    \item We propose a new algorithm for an adversarial attack to spoof robustness certificates. Unlike the prior study's constrained manipulation of an input image to seek certifiable adversarials, we consider a region-based approach to select areas for perturbation. This enables the perturbation to remain \textit{imperceptible} but more efficacious and evasive because the region selection considers \textit{natural} image boundaries and \textit{salient} regions. 

    \item  We provide a rigorous evaluation of our attack algorithm, dubbed \name{}, with the large-scale \texttt{ImageNet} task. To advance the prior evaluation of untargeted attacks, we construct \textit{targetted} attacks, including a significant effort to evaluate with a state-of-the-art certification method based on diffusion models (DensePure).

    \item We raise awareness of weaknesses in current robustness certification methods \& discuss the implications of attacks targeting both classification and certification. 
\end{itemize}

Importantly, while our new exploit demonstrates that existing certification methods can be exploited with stealth, the attack does not invalidate the certificates produced---the assertion made by a certificate that an input is not an adversarial example in the chosen bounded norm is still correct. \textit{Rather, our attack is a cautionary tale}.  
\rqq{\textbf{Key Takeaways}: Safe deployment of systems with certifiably robust models should use certificates as an indicator of label correctness with caution. A strongly certified sample input (with a large certification radius) does not necessarily imply correctness nor that a potential manipulation to spoof a certificate will lead to easily visible evidence.
We confirm, even targetted attacks (certification for a target label chosen by an attacker) are possible. But, as expected, are harder. The state-of-the-art denoiser-based method remains the most effective certification method, even when compared to model ensembling under randomised smoothing.}

We hope the study helps reveal and deepen understanding of flaws in certified defenses for adversarial robustness. 

\section{Related Work}
\label{sec:Related Work}
Adversarial attack algorithms can launch powerful attacks like Projected Gradient Descent (PGD) to craft and apply imperceptible perturbations to inputs to mislead or hijack the decision of deep learning models~\citep{Szegedy2013, papernot2017, Carlini2017, Madry2017, Athalye2018}.

The recent \shadow in~\cite{Ghiasi2020} exposes a weakness in a certified defences. The attack generates large perturbations in the input-space to move an image \textit{far} from a class boundary to a region capable of generating a fake certificate with a large radius. The attack augments PGD to constrain semantic changes and perturbations with three penalties: i)~to force the perturbation $\delta$ to have small total variation to attempt to appear smooth and natural; ii)~to limit the perturbation $\delta$ by constraining the change in the mean of each color channel; and iii)~to promote perturbations that assume similar values in each colour channel to suppress extreme or dramatic colour changes. \textit{We devise a simpler attack and demonstrate that such large perturbations are not necessary to break a certified defence}. 

\section{Preliminaries on Scalable Certification}
Complete certification methods guarantee finding adversarial examples if they exist, but are limited to small datasets and simple models due to scalability constraints~\cite{weng2018fastcomputationcertifiedrobustness}. Complete methods are typically restricted to specific architectures and struggle with large-scale tasks~\cite{cohen2019certified, hayes2020extensionslimitationsrandomizedsmoothing} like \texttt{ImageNet}~\cite{Deng2009}. Incomplete approaches, including deterministic~\citep{Lyu2020, Levine2020} and probabilistic methods, can certify the lower bound of model performance under certain $\ell_p$ norm attacks or abstain from deciding~\citep{Li2023}. \citet{lecuyer2019certifiedrobustnessadversarialexamples} introduced the incomplete method---\textit{randomized smoothing}---using Gaussian and Laplace noise. This approach offers a general and scalable approach for certification on large scale tasks such as \texttt{ImageNet}. The approach provides a non-trivial probabilistic robustness guarantee. Subsequent studies tightened the bound~\cite{cohen2019certified} and further improved certified performance by integrating adversarial training~\cite{Salman2019}, consistency regularization ~\cite{jeong2020consistency}, and model ensembling ~\cite{horvath2022boosting}.

\vspace{2mm}
\noindent\textbf{Randomized Smoothing.~}
Consider a classification problem from $\boldsymbol{x} \in \mathbb{R}^d$ to classes $\mathcal{Y}$. As introduced in \cite{cohen2019certified}, randomized smoothing is a method for constructing a new, “smoothed” classifier $g$ from an arbitrary base classifier $f$. When queried by $\boldsymbol{x}$, the smoothed classifier $g$ returns what the base classifier $f$ is most likely to return when $\boldsymbol{x}$ is perturbed by %isotropic Gaussian 
noise $\boldsymbol{\varepsilon} \sim \mathcal{N}(0, \sigma^2 I)$:
\begin{equation}
    \label{eq:randomize smoothing}
   g(\boldsymbol{x}) = \arg \max_{c \in \mathcal{Y}} \mathbb{P}(f(\boldsymbol{x} + \boldsymbol{\varepsilon}) = c)
\end{equation}

The noise level $\sigma$ is a hyperparameter of the smoothed classifier $g$ which controls a robustness/accuracy tradeoff; it does not change with the input $\boldsymbol{x}$. 

\vspace{2mm}
\noindent\textbf{Smoothed Ensemble.~}  Recent work ~\cite{horvath2022boosting} has introduced several advancements to enhance Randomized Smoothing. For a set of $k$ classifiers $\{f^l : \mathbb{R}^d \rightarrow \mathbb{R}^m\}_{l=1}^k$, a soft-ensemble $\bar{f}$ is constructed by averaging the logits $\bar{f}(x) = \frac{1}{k} \sum_{l=1}^{k} f^l(x)$, where, $f^l(x)$ are the pre-softmax outputs. With $\varepsilon \sim \mathcal{N}(0, \sigma^2 I)$, the smoothed ensemble can be formulated as follows: 
\begin{equation}
\label{eq:smoothed ensemble}
   g_e(\boldsymbol{x}) = \arg \max_{c \in \mathcal{Y}} \mathbb{P}(\bar{f}(\boldsymbol{x} + \boldsymbol{\varepsilon}) = c)
\end{equation}

\vspace{2mm}
\noindent\textbf{Denoised Smoothing.~}An alternative approach to obtaining a provably robust classifier \emph{without retraining} the underlying model has been proposed through the idea of \emph{denoised smoothing}~\cite{salman2020denoised}. Unlike prior work that primarily focus on training classifiers to withstand Gaussian perturbations---often using Gaussian noise augmentation~\cite{cohen2019certified} or adversarial training~\cite{Salman2019}---this method leaves the pre-trained classifier unchanged. Building on this idea, several recent methods have explored the use of diffusion-based denoiser such as DiffusionDenoisedSmoothing (DDS)~\cite{Carlini2022CertifiedAR}, DensePure~\cite{xiao2022densepureunderstandingdiffusionmodels} or DiffSmooth \cite{zhang2023diffsmoothcertifiablyrobustlearning}. Denoised Smoothing essentially augments the base classifier $f$ with a denoiser $D_\theta: \mathbb{R}^d \rightarrow \mathbb{R}^d$ to form a new base classifier defined as $f \circ D_\theta: \mathbb{R}^d \rightarrow \mathcal{Y}$. Assuming the denoiser $D_\theta$ is effective at removing Gaussian noise, this setup is configured to classify well under Gaussian perturbation of its inputs. Formally, with $\boldsymbol{\varepsilon} \sim \mathcal{N}(0, \sigma^2 I)$, this procedure can be defined as follow:
\begin{equation}
    g_d(\boldsymbol{x}) = \arg \max_{c \in \mathcal{Y}} \; \mathbb{P}\big[ f(D_{\theta}(\boldsymbol{x} + \boldsymbol{\varepsilon})) = c \big]
\end{equation}

\begin{figure*}[!t]
    \centering
    %\vspace{-2mm}
    \includegraphics[width=\textwidth]{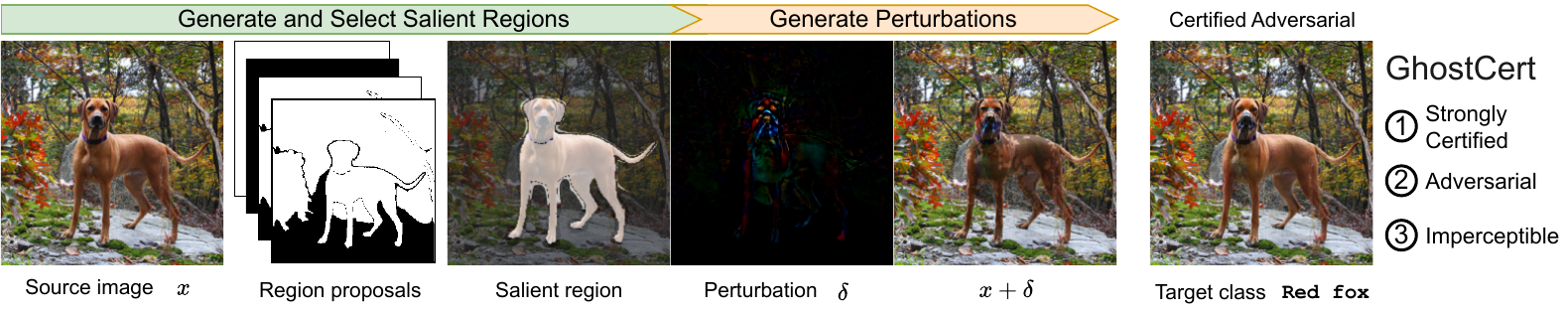}
    \vspace{-5mm}
    \caption{A pictorial illustration of \name{}. Starting from the source image $x$ with label \textbf{Rhodesian ridgeback} and given a target label \textbf{Red fox}, region proposal are evaluated to select regions for manipulation considering salient features important for classification decisions. The idea is to preserve semantics whilst minimising distortions. Then, crafting perturbations constrained to the salient regions, $\delta$, yields the adversarial $x+\delta$ misclassified as a \textbf{Red fox} while being strongly certified with imperceptible visual differences to the source image $x$.}
    \label{fig:attack_pipeline}
    \vspace{-3mm}
\end{figure*}

\vspace{2mm}
\noindent\textbf{Certifiable Robustness and Abstaining.~} 
\cite{cohen2019certified} introduces an analytic form of certifiable robustness that provides a formal guarantee for smoothed classifiers. Concretely, the prediction of a smoothed classifier remains unchanged within a bounded $\ell_2$-norm region defined by a certified radius $R$.
\begin{equation}
    R = \frac{\sigma}{2} \left[ \Phi^{-1}(p_A) - \Phi^{-1}(p_B) \right],
    \label{eq:certified radius}
\end{equation}
where $p_A$, $p_B$ are the probabilities of the top class $c_A$ and the runner-up class $c_B$ respectively, $\Phi^{-1}$ is the inverse Gaussian cumulative distribution function (CDF). Based on Theorem 1 in \cite{cohen2019certified}, the smoothed classifier $g(\boldsymbol{x})$ always returns ${c}_A$ and certified radius $\sigma\Phi^{-1}(\underline{p_A})$ if the lower bound of the probability of the top class $\underline{p_A}$ exceeds 0.5. Otherwise, the smoothed classifier abstains from making a prediction. To prevent an adversary from manipulating the smoothed classifier to abstain at a high rate, a margin $\mu$ is introduced as follows:

\begin{equation}
    \mathbb{P}(f(\boldsymbol{x}+\boldsymbol{\delta}+\boldsymbol{\varepsilon})=c_A)- {\mathbb{P}(f(\boldsymbol{x}+\boldsymbol{\delta}+\boldsymbol{\varepsilon})=c_B)}\geq \mu
\end{equation} 

\section{Proposed Method}
\label{sec:Proposed Method}
\noindent\textbf{Threat Model.~}We consider a white-box threat model for attacks on a target DNN (deep neural network), where adversaries have full access to the model’s architecture, parameters and the noise level used by the smoothed classifier. 

\vspace{2mm}
\noindent\textbf{Problem Formulation.~}Given a neural network \( f \), an input \( x \), and the ground truth label \( y \), we define an adversarial attack as an optimization problem searching for a perturbation \( \delta \) that maximizes the loss function \( L \). The objective is to generate adversarial examples misleading a smoothed classifier while ensuring the perturbation remains imperceptible. 
\begin{equation}
\begin{aligned}
    &\max_{\delta} \sum_{i=1}^{N} L(f(x + \varepsilon_i + \delta), y)~~\text{s.t. } \|\delta\|_2 \leq \epsilon,
\end{aligned}
\label{eq:single randomized smoothing}
\end{equation}
where \( L(\cdot, \cdot) \) is the loss function, \( \delta \) denotes the adversarial perturbation, \(\epsilon \) is the perturbation budget, \( \varepsilon_i \) represents isotropic Gaussian noise applied to the input, \( N \) is the number of noisy samples used to approximate the probability mass in Randomized Smoothing. The formulation of this optimization allows to find an adversarial perturbation \( \delta \) that consistently forces the smoothed model to misclassify the adversarial example $\boldsymbol{x}+\boldsymbol{\delta}$. 

When the base model is an ensemble of $k$ classifiers $\{f^l : \mathbb{R}^d \rightarrow \mathbb{R}^m\}_{l=1}^k$ and the resulting classifier is $\bar{f}$, the problem can be formulated as:
\begin{equation}
\begin{aligned}
    &\max_{\delta} \sum_{i=1}^{N} L(\bar{f}(x + \varepsilon_i + \delta), y)~~\text{s.t. } \|\delta\|_2 \leq \epsilon,
\end{aligned}
\label{eq:smoothed ensemble}
\end{equation}
\noindent Similarly, when the base model $f_\theta$ is subjected to a denoiser $D_\theta$, the new base model becomes $f \circ D_\theta$ and therefore, the problem formulation changes to:
\begin{equation}
\begin{aligned}
    \max_{\delta} \sum_{i=1}^{N} L(f(D_\theta(x + \varepsilon_i + \delta)), y)~\text{s.t. } \|\delta\|_2 \leq \epsilon.
\end{aligned}
\label{eq:denoise smoothing}
\end{equation}

\subsection{Method Intuition}
\label{sec:observartion}
To solve these attack optimization problems, several gradient-based methods such as Fast Gradient Sign Method (FGSM) \cite{Goodfellow2014}, Basic Iterative Method (BIM) \cite{Kurakin2016} can be employed. However, Projected Gradient Descent (PGD) \cite{Madry2017} has emerged and provides superior performance for white-box attacks due to its ability to navigate complex loss surfaces through iterative small steps gradient ascent. Therefore, we adopt PGD in our study to navigate toward the adversarial solution.

While traditional adversarial attacks successfully employ a global perturbation, they have not leveraged saliency information within natural images to enhance the imperceptibility of adversarial perturbation. Recent empirical observations in~\cite{rambo2022ramboattack} 
demonstrate that, when searching for perturbations to inputs in black-box settings, they tend to concentrate within salient regions of an image, although the attack \textit{does not explicitly target} these regions. This suggests that more effective perturbations could be crafted by manipulating the salient regions. 

To identify salient regions with Convolutional-based models, GradCAM \cite{selvaraju2017grad} represents a natural choice due to its widespread adoption in highlighting decision-critical areas. However, GradCAM's gradient-based saliency maps produce amorphous regions that disregard the natural structural boundaries inherent in images. This results in perturbations producing unnatural artifacts and compromising semantic coherence. To overcome this fundamental limitation, we introduce a new notion---\textit{salient-region masks}---that combines GradCAM's gradient-driven saliency information with semantic segmentation boundaries derived from the Segment Anything Model (SAM) \cite{kirillov2023segment}. Similarly, for transformer-based models, Attention maps provide an alternative to GradCAM, directly revealing which spatial locations the model prioritizes during classification. This integration ensures that perturbations maintain natural-looking boundaries while maintaining focus on salient regions. Our core hypothesis posits that constraining adversarial perturbations to these semantically-coherent salient regions will yield adversarial examples with imperceptibility and preserved semantic meaning. 

\subsection{GhostCert Attack Algorithm }
\label{sec: spoofing attack}
A pictorial illustration of how \name{} produces the adversarially perturbed image leading to misclassification and certificate spoofing while being visually imperceptible is shown in Figure~\ref{fig:attack_pipeline}.
By using standard image segmentation techniques to generate region proposals, combined with saliency analysis, to select regions defined by natural image boundaries, we aim to   generate more natural-looking adversarial examples, that are strongly certified (large spoofed certification radius, often higher or comparable with the source image). Owing to  their imperceptibility objective, we refer to these as ghost certificates and dub our method \name. 

\vspace{1mm}
\noindent\textbf{Generate Salient-Region Mask.~}
Let $\mathcal{S} \in [0,1]^{H \times W}$ be the saliency map generated by GradCAM or Attention depending on the model under attack, and let $\mathcal{\boldsymbol{M}} = \{ M_1, M_2, \ldots, M_n \}$ denote the \textit{region proposals}---realised as a set of binary segmentation masks produced by the SAM model, each with area $>300$ pixels. Additionally, let $U$ be a binary unmask candidate mask representing pixels not covered by any of the segmentation masks in $\mathcal{\boldsymbol{M}}$. For each mask $M_i \in \mathcal{M}$, the saliency overlap score is defined as:
\begin{equation}
\text{score}(M_i) = \frac{\sum_{x,y} M_i(x, y) \cdot \mathcal{S}(x, y)}{\sum_{x,y} M_i(x, y) + \sum_{x,y} \mathcal{S}(x, y)}.
\end{equation}

\noindent
Similarly, an overlap score for the unmask candidate $U$ is also obtained. Let $\mathcal{T} \subset \mathcal{M} \cup \{U\}$ be the set of top-$k$ masks selected based on the highest scores. The final combined mask---\textit{salient-region mask}---$m$ is then obtained by summing the top-$k$ masks:
\begin{equation}
m = \sum_{M \in \mathcal{T}} M.
\end{equation}

\noindent
This salient-region mask $m$ highlights the most salient and semantically meaningful regions, guided by both segmentation and GradCAM/attention-based saliency.

\vspace{1mm}
\noindent\textbf{Generate Perturbations.~}
Considering an untargeted attack setting, for every image a batch of noisy images is generated, with each image in the batch subjected to random Gaussian noise of standard deviation $\sigma$. The perturbation $\delta$, initially zero, is added to the batch of noisy images, and the following problem is optimized.

\begin{equation}
    \begin{aligned}
        &\max_{\delta} \sum_{i=1}^{N} L\left(f_{\theta}\left(x + \Delta_i + \delta \odot m\right), y\right) \\
        &\text{s.t. } \|\delta \odot m\|_2 \leq \epsilon,
    \end{aligned}
\end{equation}

\noindent where \( L(\cdot, \cdot) \) is the loss function. In this work, we use cross-entropy loss. $\odot$ denotes element-wise multiplication, \( \delta \) is the adversarial perturbation applied to the input, \( \Delta_i \) represents the random Gaussian noise of standard deviation $\sigma$, \(m\) is the selective region or mask to perturb, \(\epsilon \) is the perturbation budget. This selective perturbation, when added to the source image \(x\), while bounded by \(\epsilon \) produces an adversarial image with visually less perceptible changes. Similarly, when the attack setting is targeted, a target label \(y_{target}\) is provided instead of \(y\) since the targeted attack aims to fool the model \(f_{\theta}\) to predict the class label of image \(x\) to be \(y_{target}\). 

\begin{figure*}[htbp]
  \centering
  \setlength{\tabcolsep}{3pt}
  \renewcommand{\arraystretch}{1.0}
  \begin{tabular}{c|c c|c c}
    & \multicolumn{2}{c|}{\shadow} & \multicolumn{2}{c}{\name{} (Ours)} \\[-1.5mm]
    % ---- First Row ----
    \begin{subfigure}{0.18\textwidth}
      \caption*{Source image $x$}
      \includegraphics[width=\linewidth]{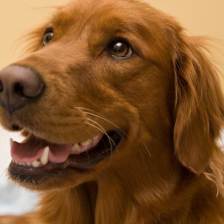}
      \caption*{\colorbox{yellow}{Radius = 0.40}}
    \end{subfigure} &
    \begin{subfigure}{0.18\textwidth}
      \caption*{Adversarial}
      \includegraphics[width=\linewidth]{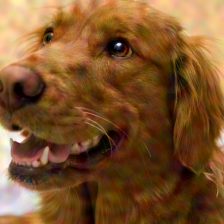}
      \caption*{\colorbox{white}{Radius = 1.14}}
    \end{subfigure} &
    \begin{subfigure}{0.18\textwidth}
      \caption*{Perturbation}
      \includegraphics[width=\linewidth]{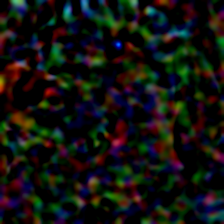}
      \caption*{\colorbox{lightgray}{Distortion = 13.28}}
    \end{subfigure} &
    \begin{subfigure}{0.18\textwidth}
      \caption*{Adversarial}
      \includegraphics[width=\linewidth]{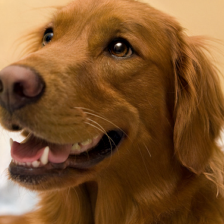}
      \caption*{\colorbox{yellow}{Radius = 1.23}}
    \end{subfigure} &
    \begin{subfigure}{0.18\textwidth}
      \caption*{Perturbation}
      \includegraphics[width=\linewidth]{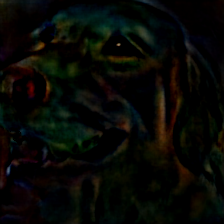}
      \caption*{\colorbox{lightgray}{Distortion = 4}}
    \end{subfigure} \\[2mm]

    % ---- Second Row ----
    \begin{subfigure}{0.18\textwidth}
      \includegraphics[width=\linewidth]{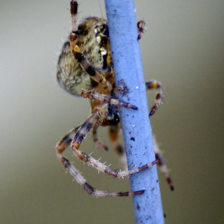}
      \caption*{\colorbox{yellow}{Radius = 0.58}}
    \end{subfigure} &
    \begin{subfigure}{0.18\textwidth}
      \includegraphics[width=\linewidth]{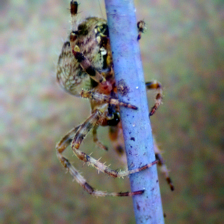}
      \caption*{\colorbox{white}{Radius = 1.23}}
    \end{subfigure} &
    \begin{subfigure}{0.18\textwidth}
      \includegraphics[width=\linewidth]{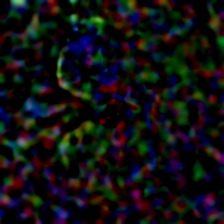}
      \caption*{\colorbox{lightgray}{Distortion = 12.94}}
    \end{subfigure} &
    \begin{subfigure}{0.18\textwidth}
      \includegraphics[width=\linewidth]{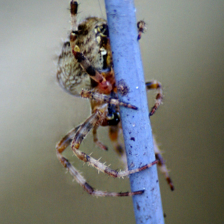}
      \caption*{\colorbox{yellow}{Radius = 1.23}}
    \end{subfigure} &
    \begin{subfigure}{0.18\textwidth}
      \includegraphics[width=\linewidth]{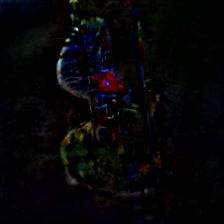}
      \caption*{\colorbox{lightgray}{Distortion = 6}}
    \end{subfigure}
  \end{tabular}

  \caption{Illustrative examples of successful attacks by \name{} are presented. For each case, we display the adversarial image and its corresponding perturbation generated by both the Shadow attack and our method, \name{}. The results clearly show that \name{} produces strongly certified adversarial examples with perturbations that are more visually imperceptible than those from the Shadow attack, while also achieving higher spoofed certification radii at lower $l_2$ norms ($||\delta||_2$).}
  \label{fig:successful_attacks}
  \vspace{-2mm}
\end{figure*}

\vspace{1mm}
\noindent\textbf{Attack pipeline.~}
We codify the attack in Algorithm 1. Figure~\ref{fig:successful_attacks} shows samples from successful attacks with \name{} and the prior \shadow{}. 

\begin{algorithm}
\caption{\name{}}
\begin{algorithmic}[1]
\Require Input image $x$, ground truth label $y$, target label $y_{\text{target}}$ (if targeted), noise $\Delta_i$, mask $m$, step size $\lambda$, maximum distortion $\epsilon$, attack type (targeted or untargeted)
\State Initialize $\delta \gets 0$
\For{$i = 1$ to $N$}
    \If{attack is targeted}
        \State $g \gets -\nabla_{\delta} L(f_{\theta}(x + \Delta_i + \delta), y_{\text{target}})$
    \Else
        \State $g \gets \nabla_{\delta} L(f_{\theta}(x + \Delta_i + \delta), y)$
    \EndIf
    \State $\delta \gets \delta + \lambda \cdot \frac{g}{\|g\|_2}$ \Comment{Gradient ascent/descent step}
    \State $\delta \gets \left( \epsilon\frac{\delta}{\|\delta\|_2} \right) \odot m$ \Comment{Projection and mask step } %to $\epsilon$-ball
\EndFor
\State \Return $\delta$
\end{algorithmic}
\end{algorithm}

\renewcommand{\arraystretch}{1.0}

\vspace{-3mm}
\section{Experiments and Evaluations}
\label{sec:Experiments and Evaluations}

\vspace{1mm}
\noindent\textbf{Dataset(s).~}We used the large-scale \texttt{ImageNet}~\cite{deng2009imagenet} validation set for experiments as in \shadow. For each base model and $\sigma$ combination, under Randomized Smoothing, 100 correctly classified images and their certification radii were identified (notably \shadow used 50 samples). This ensures the attacks were carried out considering only images that were  correctly certified by the models under Randomized Smoothing. For each set of 100 images, on average, there were 94 to 97 different class labels, suggesting a low bias towards a particular class label. 

\noindent\textbf{Defended models and attacks.~}We employed three certified defenses: i)~Randomized Smoothing (RS) with Resnet50 (Single model)~\cite{cohen2019certified}; and ii)~Ensemble of three consistency-trained Resnet50 models~\cite{horvath2022boosting} under RS (with $\boldsymbol{\sigma}={0.25, 0.5, 1.0}$); and iii)~diffusion-based denoiser prepended to a BEiT large Patch16 512 transformer under RS (DensePure) (with $\boldsymbol{\sigma}={0.25, 0.5}$)~\cite{xiao2022densepureunderstandingdiffusionmodels}. The ensemble was included as part of the defended models because it is well established that, in terms of certified robustness, an ensemble of consistency-trained models outperforms a single Resnet50, yielding higher certified accuracy under Randomized Smoothing. 

To evaluate, three attacks were carried out---\name, current state-of-the-art \shadow{}, and our modified version \shadow{} to bound the distortion limit to compare with \name{} referred to as \shadow~(bounded).

\vspace{1mm}
\noindent\textbf{Performance Metrics.~}\textbf{\textit{\underline{Attack Success Rate (ASR)}}}. In the untargeted case, it is the proportion of samples misclassified by the defending model. In the targeted case, it is the proportion of samples classified by the defending model as the target class and assigned a certificate (radius)---if the defence abstains from making a prediction, these were recorded as \textbf{\textit{\underline{Denial of Service (DoS)}}} attacks. \textbf{\textit{\underline{Spoofing Radius}}}. This is the certification radius calculated for samples that have been successfully misclassified.
     
\vspace{1mm}
\noindent\textbf{Evaluation Protocol.~}For both untargeted and targeted settings, ASR and the average Spoofing Radius were reported for evaluation across every defense method. For a targeted attack setting, due to computational constraints, only one target label was chosen for each source image. The target label was determined by sequentially searching for an image with a different label, starting from the following index of the source image. In addition, for targeted attacks, DoS results were reported as well.
This process helps eliminate bias in the selection the target label. Table~\ref{tab:eval-prtocol} summarizes the evaluation protocols used.

\begin{table}[!t]
%\begin{threeparttable}[!t]
\centering
\caption{Evaluation protocol summary.}
\label{tab:eval-prtocol}
\vspace{-3mm}
\resizebox{\columnwidth}{!}{%
\begin{tabular}{ll|l|l|l} 
\toprule
\textbf{Attack}                 & \textbf{Pert. Budgets ($\boldsymbol{\epsilon}$)} & \textbf{Defense Methods}                                                             & \textbf{Dataset}                   & \textbf{Perf. Metrics}                                                                      \\ 
\midrule
\name{}(\textbf{Ours})           & 2, 4, 6, 8, 10                                   & \multirow{3}{*}{\begin{tabular}[c]{@{}l@{}}Single\\Ensemble\\DensePure\end{tabular}} & \multirow{3}{*}{\texttt{ImageNet}} & \multirow{3}{*}{\begin{tabular}[c]{@{}l@{}}ASR\\Spoofing Radius\end{tabular}}  \\
\shadow                         & N/A                                              &                                                                                      &                                    &                                                                                             \\
Bounded \shadow$^1$ & 2, 4, 6, 8, 10                                   &                                                                                      &                                    &                                                                                             \\
\bottomrule
\end{tabular}
}
\parbox{\linewidth}{\scriptsize
Note 1: Our adapted version of Shadow Attack~\cite{Ghiasi2020} to constrain perturbations.}
\vspace{-6mm}
\end{table}
%\end{threeparttable}

\vspace{1mm}
\noindent\textbf{Attacking a (Single) and (Ensemble) Classifier Under Randomised Smoothing~}
\label{sec:attack randomized smoothing}

\begin{figure*}[!t]
  \centering
  \includegraphics[width=\textwidth]{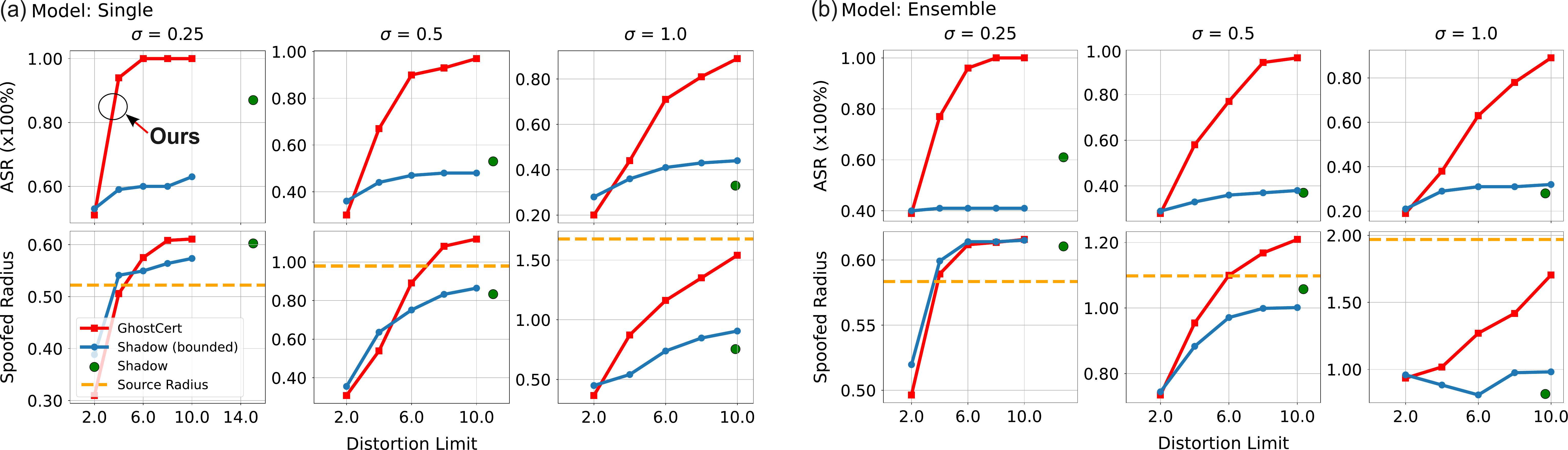}
  \caption{Comparing ASR and spoofed radii for three attacks in \textit{untargeted} settings against (a)~single ResNet-50 under Randomized Smoothing (RS) and (b) an ensemble of three consistency ResNet-50 models under RS vs. distortion  $\|\delta\|_2$ budgets.}
  \label{fig:untargeted_resnet50}
  \label{fig:untargeted_ensemble}
  \vspace{-4mm}
\end{figure*}

\begin{figure*}[!t]
  \centering
  \includegraphics[width=\textwidth]{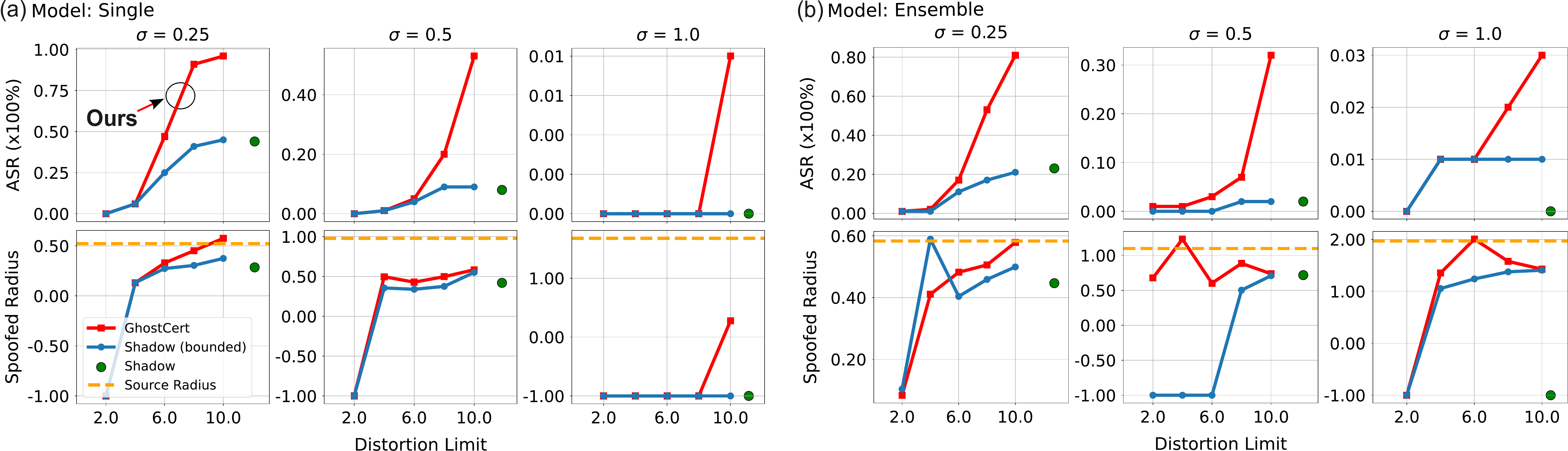}
  \caption{Comparing ASR and spoofed radii for three attacks in a \textit{targeted} setting against (a) single ResNet-50 under Randomized Smoothing (RS) and (b) an ensemble of three consistency ResNet-50 models under RS vs. distortion $\|\delta\|_2$ budgets.}
 \label{fig:targeted_resnet50}
  \label{fig:targeted_ensemble}
  \vspace{-3mm}
\end{figure*}

\vspace{1mm}
\noindent\textit{\underline{Attack Success Rate (ASR)}.~}(\textbf{\textit{Single}})~For both untargeted and targeted attacks, \name{} consistently achieves significantly higher ASR than both bounded and unbounded shadow attacks across all noise levels ($\sigma = 0.25, 0.5, 1.0$) as demonstrated in Figure~\ref{fig:untargeted_resnet50} and Figure~\ref{fig:targeted_resnet50}. 

%\vspace{1mm}
\noindent(\textbf{\textit{Ensemble}})~For untargeted attacks, at $\sigma = 0.25$ \name{} achieves $\approx100\%$ ASR while shadow attack and its variant plateau at $~40\%$ across different ($\sigma = 0.25, 0.5, 1.0$) as shown in Figure~\ref{fig:untargeted_ensemble}. The ensemble defense significantly impacts shadow attack and its variant but has minimal effect on \name{}'s effectiveness. For targeted attacks, at $\sigma = 0.25$, \name{} achieves an ASR of over $80\%$ while the shadow variants max out at just over $20\%$. At higher noise levels ($\sigma = 0.5, 1.0$), while the ASR is not as high across all attacks, \name{} still produces significantly larger ASR than the shadow attacks as shown in Figure~\ref{fig:targeted_ensemble}.

\textit{Overall, the performance gap is most pronounced at higher noise levels, where \name{} maintains effectiveness while shadow attacks degrade significantly}.

\noindent\textit{\underline{Spoofed Radii}.~}(\textbf{\textit{Single}})~For untargeted attacks, \name{} generates substantially larger spoofed radii compared to the shadow baselines, particularly evident at $\sigma = 0.5, 1.0$. Notably, \name{}'s spoofed radii consistently exceed or approach the source certified radius (\textit{orange} dashed line), indicating effective circumvention of the defense mechanism. For targeted attacks, \name{} consistently achieves larger spoofed radii than its \shadow counterparts.

\noindent(\textbf{\textit{Ensemble}})~In the untargeted setting, \name{} consistently generates larger spoofed radii than shadow baselines across all perturbation budgets. At $\sigma = 0.5$, \name{}'s spoofed radii approach or exceed the source radius (orange dashed line), indicating strongly certified defense bypass. At $\sigma = 1.0$, while all methods fall below the original radius due to stronger noise, \name{} maintains a substantial advantage. 
In a targeted setting, in cases where there is significant ASR to report, \name{} consistently generates larger spoofed radii than shadow baselines. In cases where the shadow variants generate similar spoofed radii, the ASR is negligibly low. 

\noindent\textit{\underline{DoS Success}.~}Interestingly, certification methods can abstain from making a decision. 
Failing to certify robustness for an input results in the verifier abstaining instead of making a potentially incorrect or vulnerable prediction. Table~\ref{tab:asr_abstain} reports abstain results observed in targetted attacks. When the given perturbation budget is inadequate for spoofing a certification (indicated by low ASR in Fig.~\ref{fig:targeted_resnet50}), \name{} input samples lead to a higher DoS success. Importantly, when the ASR for \name{} is similar to \shadow, the DoS success for our attack is generally higher than both shadow variants. This indicates the region-based perturbations are more effective but the adversarial crafted is near a decision boundary and the $\epsilon$-bound it too large to spoof a certificate.

\renewcommand{\arraystretch}{1}
\setlength{\tabcolsep}{3pt}
\begin{table}[t!]
\vspace{-1mm}
\centering
\caption{DoS (Abstain) attack success (\%) (Single).}
\vspace{-2mm}
\resizebox{0.78\columnwidth}{!}{%
\begin{tabular}{c c c c c c c c c c}
    \toprule
    \textbf{$||\boldsymbol{\delta}||_2$}
    & \multicolumn{3}{c}{\textbf{Shadow ($\sigma$)}} 
    & \multicolumn{3}{c}{\textbf{Shadow (Bounded) ($\sigma$)}} 
    & \multicolumn{3}{c}{\textbf{\name{} ($\sigma$)}}
    \\
    \cmidrule(lr){2-4}
    \cmidrule(lr){5-7}
    \cmidrule(lr){8-10}
    & $0.25$ & $0.5$ & $1.0$
    & $0.25$ & $0.5$ & $1.0$
    & $0.25$ & $0.5$ & $1.0$
    \\
    \midrule
    2 & \multirow{5}{*}{25} & \multirow{5}{*}{27} & \multirow{5}{*}{34}
      & 14 & 8 & 9 & 8 & 84 & 8 \\
    4 &  &  &  & 31 & 18 & 18 & 20 & 68 & 12 \\
    6 &  &  &  & 25 & 29 & 23 & 16 & 52 & 21 \\
    8 &  &  &  & 24 & 29 & 35 & 4 & 38 & 34 \\
    10 & &  &  & 21 & 38 & 39 & 2 & 45 & 45 \\
    \bottomrule
\end{tabular}
}
\label{tab:asr_abstain}
\end{table}
\renewcommand{\arraystretch}{1}

\begin{figure}[t!]
  \centering
  \includegraphics[width=0.88\linewidth]{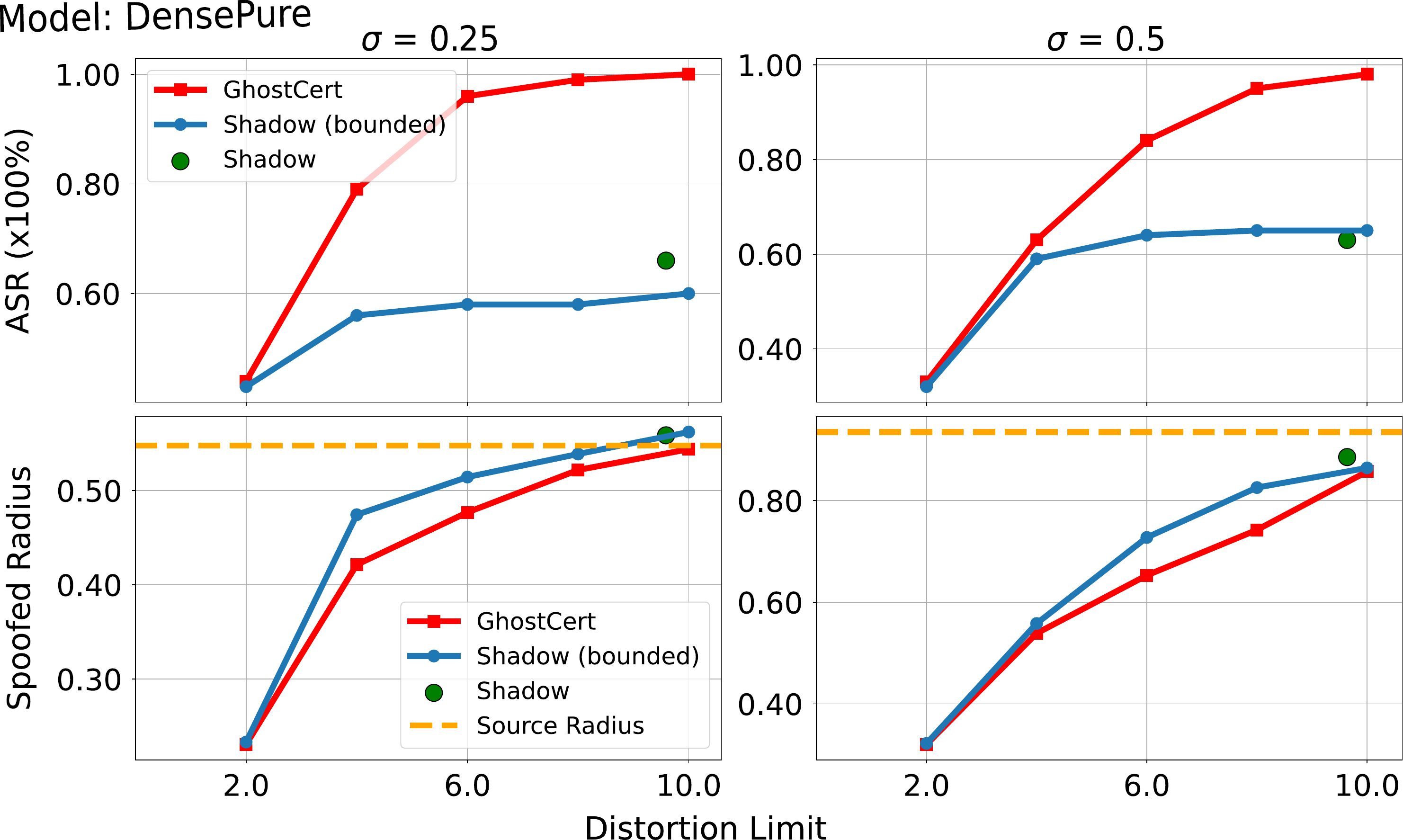}
  \caption{Comparing ASR and spoofed radii between three attacks in an \textit{untargeted} setting against DensePure.}
  \label{fig:untargeted_densepure}
  \vspace{-2mm}
\end{figure}

\vspace{1mm}
\noindent\textbf{Attacking Denoised Smoothing (DensePure)}\\

\noindent\textit{\underline{Attack Success Rate (ASR)}.~}\name{} demonstrates superior and consistent performance across both noise levels as shown in Figure~\ref{fig:untargeted_densepure}. At ($\sigma = 0.25$) \name{} significantly outperforms shadow attacks and its variant across different perturbation budgets. \name{} maintains $30-100\%$ success rates while shadow method and its variant remain constrained at $30-65\%$. \textit{\underline{Spoofed Radii}.~}Across pertubations budgets, \name{} achieves spoofed certification radii that are slightly lower or comparable with \shadow{} but consistently maintains a significantly higher ASR.

\begin{figure}[t!]
    \centering
    \vspace{-1mm}
    \includegraphics[width=\columnwidth]{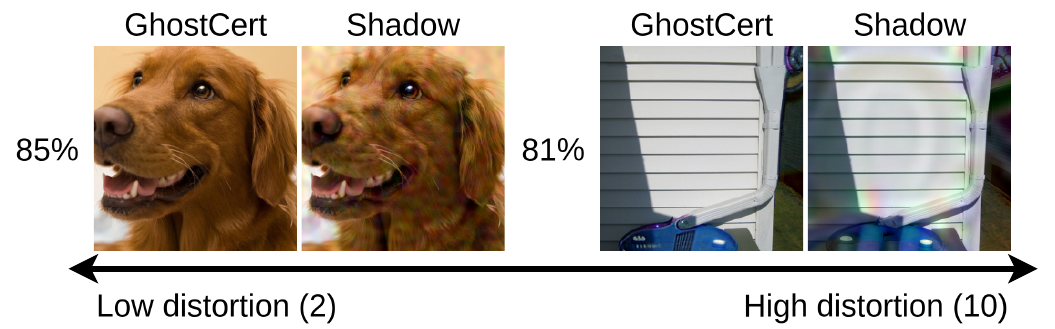}
    % \vspace{2mm}
    % \small
    \begin{minipage}{\columnwidth}
    \footnotesize
        \vspace{1.5mm}
        \centering
        \begin{tabular}{l|l|l|l|l|l}
        \toprule
        Distortion budget $\|\delta\|_2$ & 2 & 4 & 6 & 8 & 10 \\
        \midrule
        \textbf{Avg. \% Selecting \name{}} & 74 & 69 & 58 & 60 & 62 \\
        \bottomrule
        \end{tabular}
        %\vspace{1pt}
        \scriptsize
        \makecell[l]{
        \textit{\textbf{~~~~~~~~~~~~~Additional images and full results are in the Appendix.}}} 
    \end{minipage}
    \caption{Naturalism/imperceptibility of the adversarial images generated by \name{} vs. \shadow{} across minimum \& maximum distortion budgets. \name{} images were consistently perceived as \textit{more} natural looking.}
    \label{fig:naturalism_score}
    \vspace{-4mm}
\end{figure}

\vspace{1mm}
\noindent\textbf{Evaluation of Imperceptibility: User Study}\\
A user study compare the perceptual realism of adversarial images generated by \name{} and \shadow{}. For each distortion level ($\|\delta\|_2$), 10 successful image pairs (1 as control) were presented to workers on Amazon Mechanical Turk, with the display order of images from \shadow{} and \name{} randomized. To reduce noise, responses from workers who answered randomly or spent less than a minute on the task were excluded. The results from participants for each distortion level are shown in Figure~\ref{fig:naturalism_score}. \name{} consistently produced images rated as more natural across both high/low distortion levels.

%%%%%%%%%%%%%%%%%%%%%%%%%%%%%%%

\section{Conclusion}
We show region-based input manipulations preserve semantics while subtly shifting inputs  to cause misclassification and yet receive large-radius certificates. Our method, \name{}, outperforms the state-of-the-art \shadow{} attack by achieving higher success in both misclassification and certificate spoofing, while producing more natural-looking, imperceptible adversarials. Our findings urge caution in using certification frameworks and encourage further research into certification methods and attack vectors.

\appendix
% !TEX root = ./__main.tex
\clearpage

\section{Overview of Materials in the Appendices}
\label{apd:overview}

\noindent{We provide a brief overview of the set of additional experimental results and findings in the Appendices (Please see the extended version on \url{https://arxiv.org/.})}

\begin{enumerate}
%\begin{itemize}
    \setlength{\itemsep}{2px}
    \item A detailed analysis of our attack.
    \item Details on metrics, hyper parameters and datasets.
    \item Ablation studies assessing how various mask construction methods affect ASR in both targeted and untargeted attack scenarios.
    \item Extended user study results comparing the adversarial images from \name{} and the prior SoTA method \shadow{}. %
    \item Investigating DoS success and attack success rate under targetted attack settings. 
\end{enumerate}

\bibliography{aaai2026}

\section{Attack Analysis}
\label{appdx: attack analysis}
In this section, we first articulate the novelty of the proposed attack and delineate its fundamental distinctions from conventional approaches---e.g., PGD~\citep{Madry2017}, Wasserstein~\citep{Levine2020}, and semantic attacks~\citep{Shahin2020, Bhattad2020}--with particular emphasis on differences in the threat model and optimization objective. We subsequently analyze the effectiveness of the proposed method compare to \shadow. 

\vspace{1mm}
\noindent\textbf{A new threat model.~}Standard attacks (including $\ell_p$-, Wasserstein- and semantic-based methods) are designed to induce misclassification of the base classifier $f$ while keeping perturbations imperceptible.  Our goal is strictly harder: we target certified defenses---\textit{randomized smoothing}---and aim to produce inputs that are misclassified by the smoothed classifier $g$ and simultaneously receive a large, yet spurious, certified radius (not just the base classifier $f$). Attacking a certifier requires controlling the classifier’s behavior across a neighborhood of noisy samples, not just at a single point.

\vspace{1mm}
\noindent\textbf{New objective.~}PGD is designed to force misclassification of the base classifier. Our objective is to simultaneously induce misclassification and manipulate the randomized-smoothing certifier, producing adversarial inputs that are both incorrect and receive a (spoofed) robustness certificate. To achieve this goal, we have to maximize sample agreement under the smoothing distribution rather than a single sample. This is reflected in the certification-aware losses and constraints in \eqref{eq:single randomized smoothing}, \ref{eq:smoothed ensemble} and \ref{eq:denoise smoothing}. Further,  in the targeted setting we investigate, we observe a potential new threat surface, Denial of Service attacks, against certified models not explored in \shadow.

\vspace{1mm}
\noindent\textbf{Attack Effectiveness and comparison with \shadow{}.~}The \shadow focuses on generating visually natural and smooth perturbations by regularizing pixel-level changes through total variation and other constraints rather than imperceptibility. Its optimization goal is to ensure perturbations look natural, while still crossing the classifier’s decision boundary. Thus, it turns semantic constraints into regularizers and adds them to the attack loss. This creates a multi-objective optimization where the optimizer must trade off attack strength against perceptual/semantic penalties, which can make it difficult to find low-distortion solutions that also satisfy certification criteria. We take another approach to maintain natural-looking but still achieve low distortion by formulating the problem as a constrained optimization. We restrict the perturbation in a salient and natural object bounded area by using semantic segmentation and GradCAM-derived masks. This reduces the search space and focuses the optimization on semantically coherent regions. Therefore, it improves the effectiveness our attack and achieve significantly low distortion.

\section{Task Definitions and Experiment Setup}

\noindent\textbf{Metrics.~}We provide more formal descriptions of the metrics used in our evaluations below for completeness.

\begin{itemize}
\item\textbf{Attack Success Rate (Untargeted settings)}: We follow the definition in \shadow{} and ASR is calculated by
\begin{align*}
    \left\{ \frac{\# \text{of samples not certified as souce label}}{\text{Total number of attack samples}}\right\}.
\end{align*}

%\vspace{2mm}
\item\textbf{Attack Success Rate (Targeted settings)}: This defines the success of rate of spoofing a certificate targeted attack. We defined the metric as follows:
    \begin{equation*}
        \left\{ \frac{\# \text{of samples certified as the target class}}{\text{Total number of attack samples}}\right\}.
    \end{equation*}

%\vspace{2mm}
\item\textbf{Denial of Service (DoS)/Abstain Success Rate (Targeted settings}: We defined the metric as follows:
    \begin{equation*}
        \left\{ \frac{\# \text{of samples resulting in an abstain decision}}{\text{Total number of attack samples}}\right\}.
    \end{equation*}

%\vspace{2mm}
\item\noindent\textbf{Score}: Used to measure naturalism/imperceptibility with human annotators by counting votes as follows:
\begin{equation*}
        \left\{ \frac{\# \text{times \name{} image is selected}}{\text{The total number of human annotator participants}}\right\}.
    \end{equation*}
    We expect scores to be higher for \name{} as the perturbations are less perceptible and more annotators in Amazon Mechanical Turk (AMT) agree with the same answer. 

\end{itemize}

\vspace{2mm}
%\begin{itemize}
\noindent\textbf{Dataset.~}We employed ImageNet (ILSVRC2012), which is a subset of the ImageNet dataset specifically used for the ImageNet Large Scale Visual Recognition Challenge in $2012$ containing over $1.2$ million images  distributed across $1,000$ different classes.
%\end{itemize}

\vspace{2mm}
\noindent\textbf{Hyper-Parameters.~}Detailed information about the hyper-parameters used can be found in Table~\ref{tab:hyper-parameters}.
\begin{table}[t]
\caption{\small{Hyper-parameters setting in our experiments.}}
\centering
\resizebox{.9\columnwidth}{!}{%
\begin{tabular}{ccc}
\hline
Name  & Value & Notes                                \\ \midrule
$\sigma$     & 0.25, 0.5 and 1.0     & Standard deviations of isotropic gaussian noise                          \\
$||\delta||_2$ &   2, 4, 6, 8, 10    & Maximum distortion budgets for perturbation \\
$\lambda$ & 0.0001 & The step size for gradient descent \\
$\alpha$ & 0.001 & Failure probability in Randomized Smoothing\\
$N_0$ & 10 & The number of Monte Carlo samples to use for selection\\
$N$ & 1000 & The number of Monte Carlo samples to use for estimation\\
$k$ & 5 & The number of masks used to generate the final mask after saliency analysis\\
\hline
\end{tabular}%
}
% \hline
\label{tab:hyper-parameters}
\end{table}

\section{Ablation Study}
We conducted several ablation studies to examine how different mask construction strategies influence ASR under both targeted and untargeted attack settings. Specifically, we evaluated:
\begin{itemize}
    \item A comparison between our saliency-guided region selection approach and a random-region baseline, where each pixel is independently selected with 50\% probability. The results, shown in Table~\ref{tab:ablation_r} show that our method achieves higher ASR, particularly under the harder, targeted attacks.
    \item A comparison between our saliency-guided region selection and using $k$ randomly chosen regions (without saliency). In both cases, $k$ is set to 5 since we selected the top 5 in our approach. The distinction is in region selection: our method selects the top $k$ regions from segmentation based on saliency ranking, while the comparison baseline selects $k$ regions uniformly at random. As shown in Table~\ref{tab:ablation_r_sam}, ours consistently achieves higher ASR in both targeted and untargeted settings.
    \item The sensitivity of ASR with respect to the choice of $k$---which determines the number of regions, ranked by saliency,  aggregated to construct the region for perturbation or the \textit{salient region}. As shown in Table~\ref{tab:ablation_k}, when $k = 3$, the ASR is lower in both targeted and untargeted settings. Setting $k$ to 5 or 7 leads to similar ASR across both targeted and untargeted settings. In our approach, we chose the top 5 masks from SAM based on saliency analysis.
\end{itemize}

\begin{table}[t]
\centering
\small
\renewcommand{\arraystretch}{1.3}
\begin{tabular}{c c c c c}
\hline
\textbf{Defense} & \textbf{$\epsilon$} & \textbf{Attack} & \textbf{ASR (Random)} & \textbf{ASR (Ours)} \\
\hline
\multirow{4}{*}{\makecell{Ensemble \\ ($\sigma = 0.5$)}} 
  & 2  & \multirow{2}{*}{Untargeted} & 33.33 & 35 \\
  & 10 &                         & 90 & 90 \\
  & 2  & \multirow{2}{*}{Targeted}   & 0 & 0 \\
  & 10 &                         & 5 & 30 \\
\hline
\end{tabular}
\caption{ASR comparison between region-based and  saliency guided perturbations--\name{}---and a 50\% random region baseline for 20 images from  our \texttt{ImageNet} test set.}
\label{tab:ablation_r}
\end{table}

\begin{table}[t]
\centering
\small
\setlength{\tabcolsep}{3pt}
\renewcommand{\arraystretch}{1.3}

\begin{tabular}{c c c c c}
\hline
\textbf{Defense} & \textbf{$\epsilon$} & \textbf{Attack} & \makecell{\textbf{Region}\\\textbf{Proposals only}} & \makecell{\textbf{Region Proposals}\\\textbf{+ Saliency (ours)}} \\
\hline
\multirow{4}{*}{\makecell{Ensemble \\ ($\sigma = 0.5$)}} 
  & 2  & \multirow{2}{*}{Untargeted} & 10 & 35 \\
  & 10 &                            & 45 & 90 \\
  & 2  & \multirow{2}{*}{Targeted}   & 0  & 0  \\
  & 10 &                            & 0  & 30 \\
\hline
\end{tabular}

\caption{ASR comparison between region-based and saliency guided perturbations (ours) and $k$ random regions (binary segmentation
masks or region proposals produced and selected without saliency guiding). Our method selects the top $k=5$ region proposals ranked by saliency, while the baseline selects $k=5$ random regions. Results show higher ASR for our method in both (targeted and untargeted) attack settings for 20 images from  our \texttt{ImageNet} test set.}
\label{tab:ablation_r_sam}

\end{table}

\begin{table}[t!]
\centering
\small
\setlength{\tabcolsep}{3pt}
\renewcommand{\arraystretch}{1.3}

\begin{tabular}{c c c c c c}
\hline
\textbf{Defense} & \textbf{$\epsilon$} & \textbf{Attack} 
& \makecell{\textbf{ASR,}\\\textbf{$k=3$}}
& \makecell{\textbf{ASR,}\\\textbf{$k=5$}}
& \makecell{\textbf{ASR,}\\\textbf{$k=7$}} \\
\hline
\multirow{4}{*}{\makecell{Ensemble \\ ($\sigma = 0.5$)}} 
  & 2  & \multirow{2}{*}{Untargeted} & 25 & 35 & 30 \\
  & 10 &                            & 90 & 90 & 90 \\
  & 2  & \multirow{2}{*}{Targeted}   & 0  & 0  & 0  \\
  & 10 &                            & 15 & 30 & 35 \\
\hline
\end{tabular}

\caption{Sensitivity Analysis. ASR achieved with difference choice of $k$ used to generated the salient region (see Figure~\ref{fig:attack_illustration}). Using $k=3$ results in noticeably lower ASR for both targeted and untargeted attacks, whereas $k=5$ and $k=7$ yield comparable performance. We used $k=5$ for all our evaluations.}
\label{tab:ablation_k}

\end{table}

%%%%%%%%%%% User study %%%%%%%%%%%%%%%%%%%%%%%%%%
\section{User study}
\label{apd:user-study}
We conducted a user study to evaluate the perceptual realism of adversarial images generated by \name{} and \shadow{}. For each distortion level ($|\delta|_2$), 10 successful adversarial image pairs were shown to human annotators on Amazon Mechanical Turk (AMT), with the image order between \shadow{} and \name{} randomized to remove bias. To ensure high-quality responses, participants who answered randomly or completed the task in under a minute were excluded. We determined random answers by placing an image among the 10 with very obvious deviations and manipulation in the set. Subsequently we used responses to the pair with this test image to remove responses that were unreliable.

\begin{figure}[t!]
  \centering
  \includegraphics[width=\linewidth]{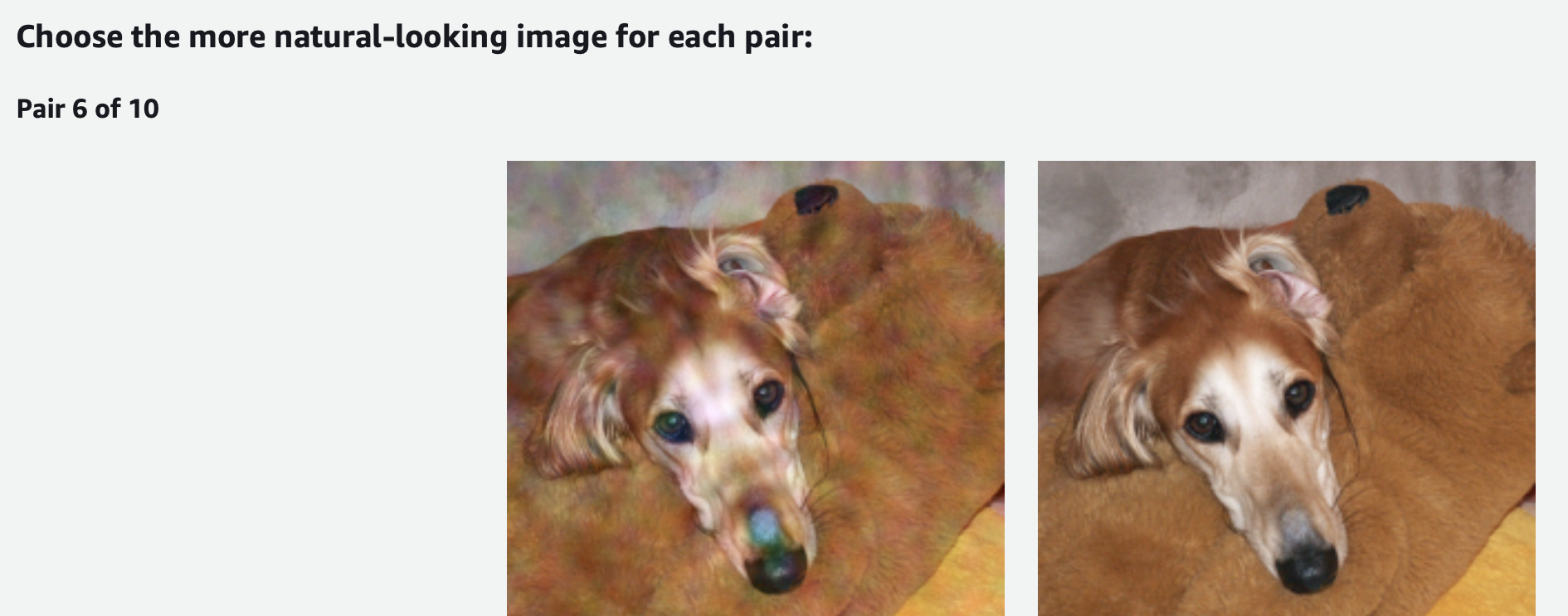} \\[2ex]
  \caption{The UI presented to AMT annotators. Once an option is chosen,  the users are presented with the next image pair.
  }
  \label{fig:survey_ui}
\end{figure}

\begin{figure*}[t!]
  \centering
  \includegraphics[width=1.0\linewidth]{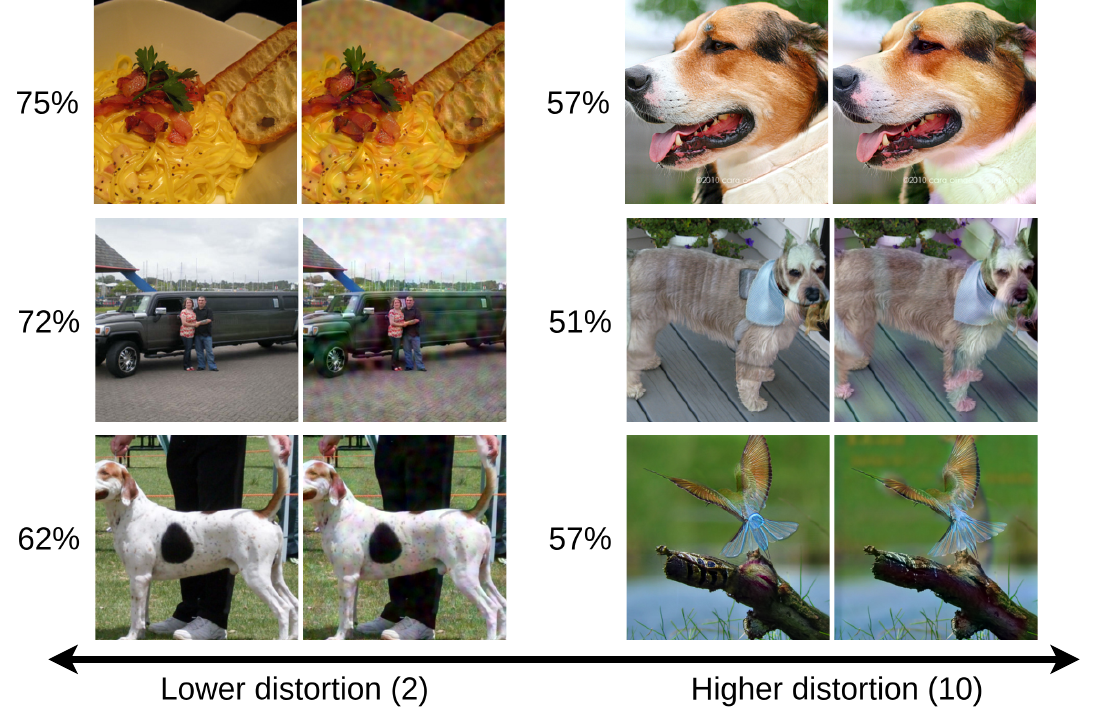}
  \caption{Naturalism (imperceptibility) of adversarial images generated by \name{} and \shadow{} across minimum and maximum distortion levels. Images produced by \name{} were consistently rated as more natural. At a higher distortion level ($\|\delta\|_2 = 10$), as expected, the difference in perceived naturalness was less pronounced, although \name{} images were still considered to be more natural looking.}
  \label{fig:naturalism_app}
  % \vspace{-6mm}
\end{figure*}

The study interface was paginated, preventing users from anticipating upcoming images. Figure~\ref{fig:survey_ui} shows the UI that is presented to an annotator in AMT. Figure~\ref{fig:naturalism_app} shows results from nearly 50 valid participants for distortion levels 2 and 10, revealing that \name{} consistently produced images perceived as more natural across both low and high distortion levels.

\begin{figure*}[t!]
  \centering
  \includegraphics[width=1.0\linewidth]{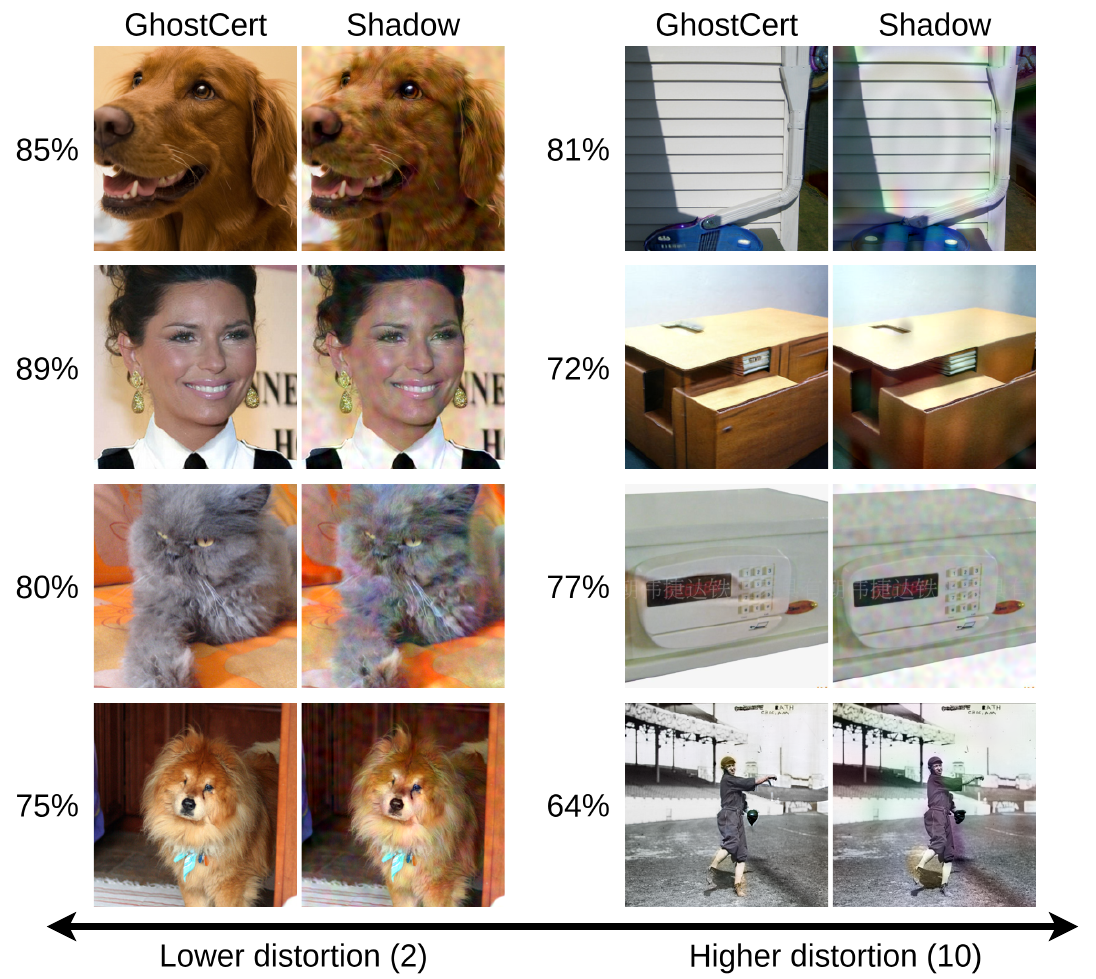}
\end{figure*}
\clearpage
\clearpage

\onecolumn 
\section{Investigating DoS Success and Attack Success Rate}

\begin{figure*}[h!]
  \centering
  \includegraphics[width=\linewidth]{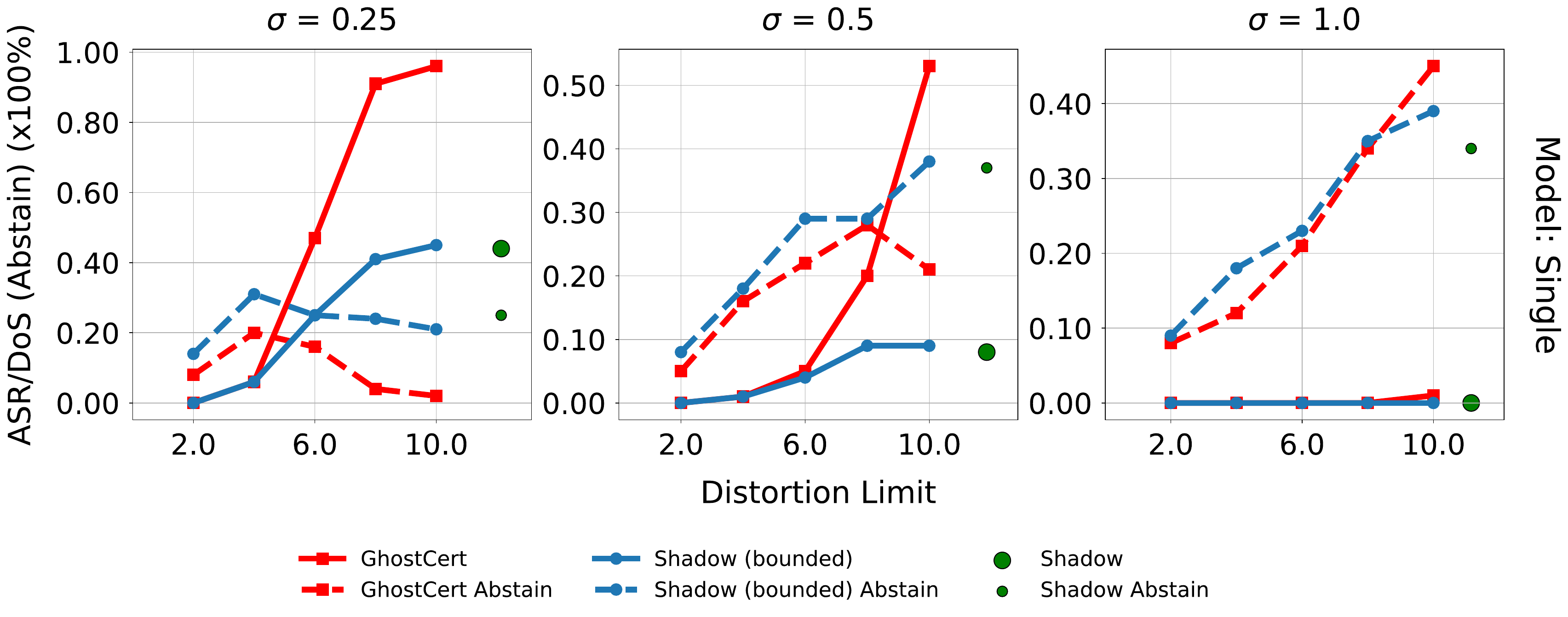}
  \caption{Comparing of DoS (abstain) and ASR (misclassified with the verifier issuing a certification radius) between three attacks in a targeted setting against a Single model (ResNet-50) under RS. As expected when ASR is high, the DoS rate is low, while ASR of \name{} is higher, generating spoofed certificates under a budget of 10 is hard for both methods for model trained with $\sigma=1.0$.}
  \label{fig:targeted_single_abstain}
  % \vspace{-6mm}
\end{figure*}

\begin{figure*}[h!]
  \centering
  \includegraphics[width=\linewidth]{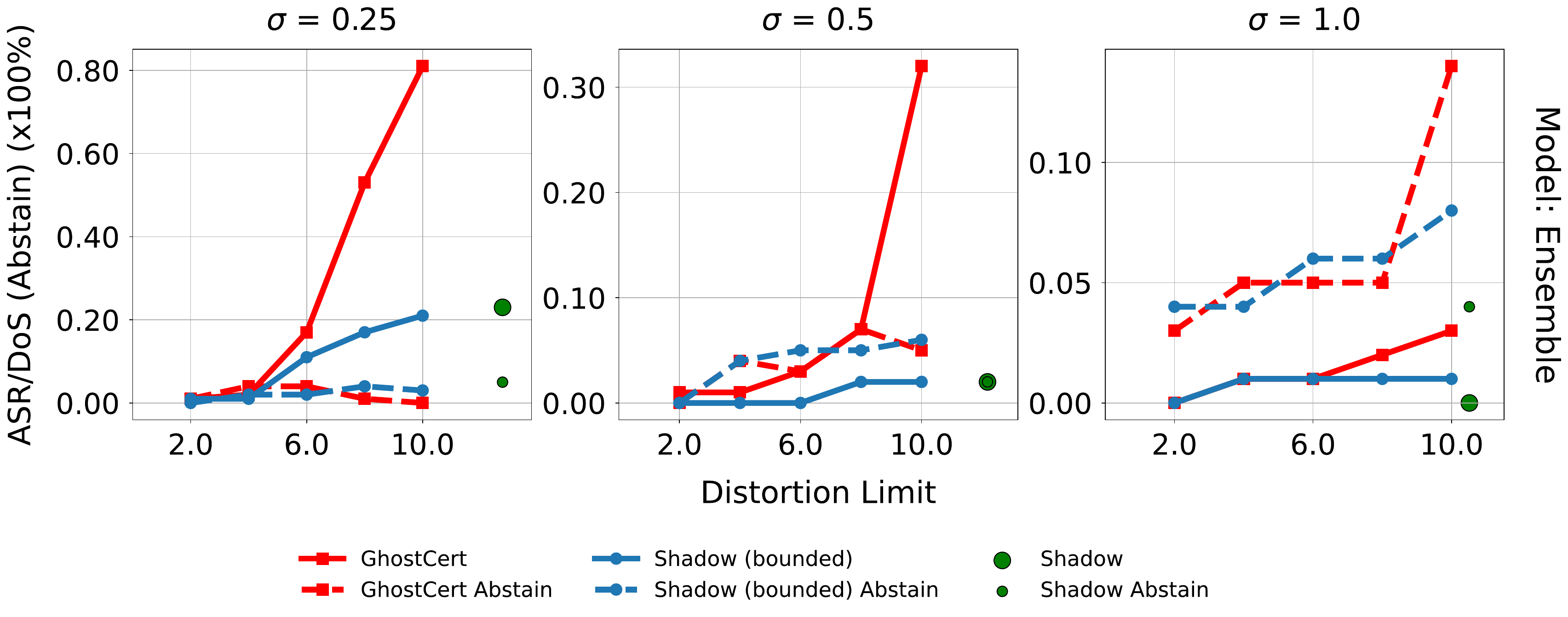}
  \caption{Comparing of DoS (abstain) and ASR (misclassified with the verifier issuing a certification radius) between three attacks in a targeted setting against an ensemble of three consistency ResNet-50 models under RS. As expected when ASR is high, the DoS rate is low, while ASR of \name{} is higher, generating spoofed certificates under a budget of 10 is is increasingly harder for both methods for model trained with $\sigma=0.5$ and 1.0. Interestingly, \name{} is observed to produce more effective adversarial against more robust models as the DoS success is higher against the $\sigma=0.1$ ensembler under RS.}
  \label{fig:targeted_ensemble_abstain}
  % \vspace{-6mm}
\end{figure*}

\end{document}